# Shape of Elephant: Study of Macro Properties of Word Embeddings Spaces

**Alexey Tikhonov**
Yandex
Berlin, Germany

## Abstract

Pre-trained word representations became a key component in many NLP tasks. However, the global geometry of the word embeddings remains poorly understood. In this paper, we demonstrate that a typical word embeddings cloud is shaped as a high-dimensional simplex with interpretable vertices and propose a simple yet effective method for enumeration of these vertices. We show that the proposed method can detect and describe vertices of the simplex for GloVe and fasttext spaces.

## Intro and related works

Neural networks for language processing have advanced rapidly in recent years. In particular, pre-trained word representations became a key component in many neural language understanding models. Word embeddings generated by neural network methods such as word2vec (Mikolov et al. 2013), glove (Pennington, Socher, and Manning 2014), fasttext (Mikolov et al. 2018) are well known to unsupervisedly exhibit linear analogical behavior (Mikolov, Yih, and Zweig 2013), (Levy and Goldberg 2014).

Although such linguistic regularity suggests the existence of some global order, interpretable directions, and the regular shape of the embeddings cloud, however, nor PCA neither t-SNE doesn't reveal any clear macrostructure of the embeddings space (Figure 1, left). Typically, one uses a dataset with labeled words to guide a supervised search for directions and transformations of the embeddings space, meaningful in terms of sentiment analysis (Yu et al. 2017), semantic categories (Senel et al. 2018), (Hennigen, Williams, and Cotterell 2020), or even the sound symbolism (Yamshchikov, Shibaev, and Tikhonov 2019).

Some works suggest that the embedding process should lead to a convex shape of the cloud, i.e. an n-dimensional simplex (Demeter, Kimmel, and Downey 2020). However, the global geometry of the word embeddings remains poorly understood and unsupervised analysis of the structure of embeddings is still an open field of research.

In this paper, we demonstrate that a typical word embeddings cloud is shaped as a high-dimensional simplex with interpretable vertices and propose a simple yet effective method for enumeration of these vertices. We show that the proposed method can detect and describe vertices of the simplex for GloVe and fasttext spaces.

## Approach

Our method relies on the following intuition: let's assume that the points in the embedding space fill a convex n-dimensional polyhedron. If the density of distribution of points in space is regular enough, then the axes found by the PCA algorithm for the point cloud should be parallel to the axes connecting some corners of the polyhedron. This is obviously true at least for the case of the uniform density and the case when the density is decreasing as the distance to the geometric center of the cloud is growing. Instead of proving the validity of these assumptions, we search the vertices of the polytope and then check the found polytope for convexity.

Now, if the axis found by the PCA is parallel to the axis connecting which two vertices (corners) of the polytope, then in the projection onto this axis, the extreme points on both ends will be points that located in the corners of the polytope. Each PCA axis gives us two vertices, while we can expect that the first found vertices will be the most contrasting in some sense. The found vertices then need to be cleaned from repetitions and false vertices. At the stage of cleaning, the algorithm compares the found vertices by top K nearest words and glues vertices that are similar above a certain threshold, and then for each vertex, it checks what percentage of words lies outside the triangle formed by this vertex with two other randomly selected ones. If the percentage of words outside the triangle is high, then the point is not a simplex vertex.

Figure 1 (right) and Figure 2 show examples of projections onto the plane formed by several of the triplets of vertices for the fasttext and GloVe spaces. It's clear that:

- the majority of the cloud's projection onto such a plane lies inside the triangle formed by vertices, i.e. the cloud appears to be an n-dimensional simplex;
- the algorithm successfully finds the simplex vertices;
- top K words of vertices are usually clearly interpretable.

## Results and Conclusion

All experiments were carried out on publicly available GloVe and fasttext word embeddings. In both cases, we used spaces with 300 dimensions and top 50K words by frequency. In the case of both datasets, the method was able to find the simplex vertices. The projections on the plane formed by the triplets of these angles (Figure 2) demonstrate an understandable macrostructure of space, in contrast to the standard PCA and t-SNE methods (Figure 1). For any triple of vertices on average 98% of the words are located inside the triangle of these vertices while projected on the plane of this triple (and 15% are located outside the incircle).

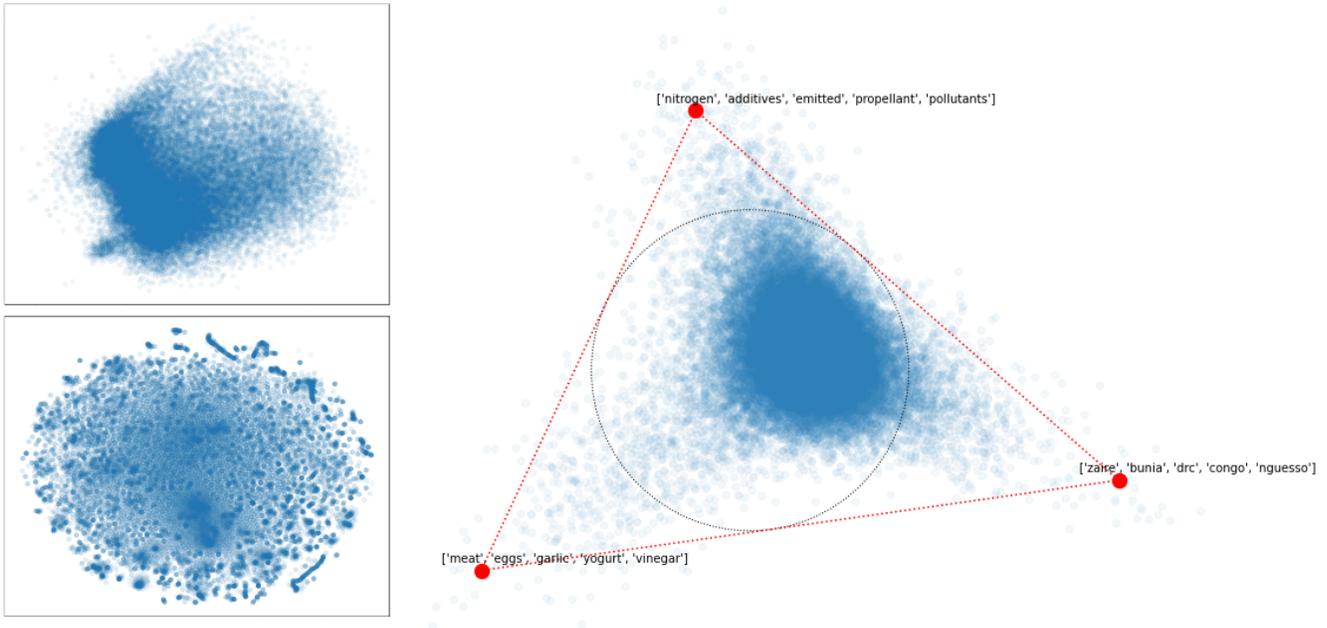

Figure 1: Projections of the same word embeddings space (fasttext): PCA (upper-left), t-SNE (bottom-left), our method (right).

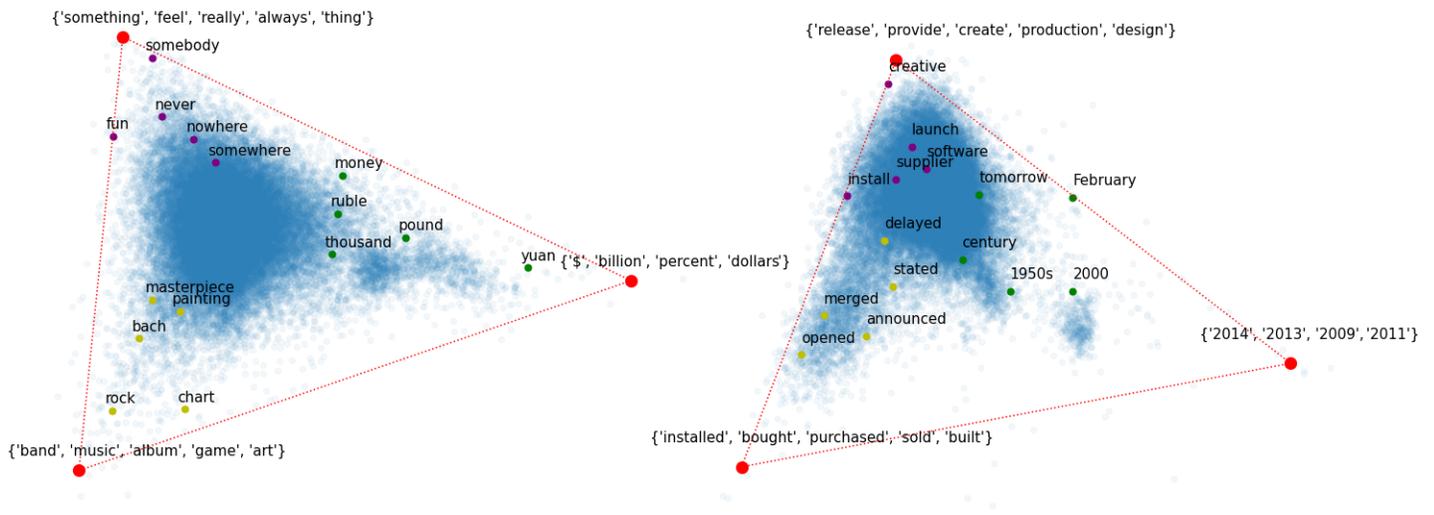

Figure 2: Examples of projections by our method for different word embeddings spaces: glove (left) and fasttext (right).

The found vertices are usually well described by the top 5 of their words. In both spaces, among the first vertices, there were ones corresponding to dates, economics, toponyms, food, common names, religion, sport, art, etc. On the other hand, artifacts specific to the training dataset are also detected – parts of markup syntax, elements of the pronunciation alphabet, camel-cased words from news titles, etc. Figure 2 also demonstrates how cosine distance to such vertices can be used for the interpretation of other words.

Our experiments show that the word embeddings cloud has the shape of the n-dimensional simplex with the dense center and a bunch of interpretable vertices. We present the method to find and describe the vertices of this simplex. The adaptation of the proposed method for the modern text embeddings spaces analysis is an item of future work.